\documentclass[conference]{IEEEtran}

\usepackage[T1]{fontenc}
\usepackage[utf8]{inputenc}
\usepackage{graphicx}
\usepackage{booktabs}
\usepackage{amsmath}
\usepackage{url}
\usepackage{listings}
\usepackage[hidelinks]{hyperref}
\hypersetup{pdftitle={Agreeing on the Answer but Not the Reason: Measurement
Limits of LLM Reasoning Failure Taxonomies}, pdfauthor={Madhu Sudhan Rao Beesetty},
pdfsubject={IEEE CogMI 2026},
pdfkeywords={evaluation methodology, LLM-as-judge, inter-rater reliability, construct validity, reasoning failure analysis, multilingual reasoning}}

\lstdefinestyle{prompt}{
  basicstyle=\fontsize{6.8}{8}\selectfont\ttfamily,
  breaklines=true,
  breakatwhitespace=true,
  columns=fullflexible,
  keepspaces=true,
  showstringspaces=false,
  frame=single,
  framesep=2pt,
  xleftmargin=1pt,
  xrightmargin=1pt,
  aboveskip=4pt,
  belowskip=4pt,
}

\begin{document}

\title{Agreeing on the Answer but Not the Reason:\\Measurement Limits of
LLM Reasoning Failure Taxonomies}

\author{\IEEEauthorblockN{Madhu Sudhan Rao Beesetty}
\IEEEauthorblockA{\textit{PayPal, Inc.}\\
San Jose, CA, USA\\
madhusudhanbeesetty@gmail.com}}

\maketitle

% ---------------------------------------------------------------------
\begin{abstract}
Failure taxonomies for LLM reasoning are increasingly reported as
measurements: papers state what fraction of a model's errors fall into named
categories, then compare those fractions across models. We show that for one
such category the reported fraction is largely a property of the measurement
procedure. We defined \emph{Compositional Collapse} (a model notices a
required structural sub-part of a problem, then drops or misapplies it) and
measured it across twelve models and eleven languages on a public multilingual
math benchmark (2{,}316 graded records). Three findings follow. First,
50.1\,\% of failures carry evidence for more than one category, so the
reported plurality is set by the order in which the classifier tests them:
reversing two tests moves it in eleven of twelve models, and across all six
effective orderings Collapse's share of failures spans 26.4\,\% to 76.4\,\%,
winning under exactly half. Second, agreement collapses for the category but
not for the task. Native speakers of all eleven languages judged 179
transcripts, agreeing with our inspector at $\kappa = 0.932$ on answer
correctness and $\kappa = 0.093$ on Collapse. A second independent native
speaker per language re-rated 142 of those items; on the 121 pairs with
determinate labels the two humans agree with \emph{each other} at
$\kappa = 0.659$ while each agrees with the inspector at $\kappa \leq 0.094$
(21 second-round responses were ``Unclear'' and are analysed separately). The
category is therefore reproducible between humans on determinate cases under a
shared definition and example, but is not reproduced by the original inspector: an
instrument-validation failure rather than an intrinsically unusable construct.
Re-running the identical inspector
model (Claude Sonnet~5) with the human instructions verbatim --- same narrower
definition, same example, same binary question --- raises $\kappa$ against
Human~2 from 0.147 to 0.636 on the 106 of those pairs that it labels
determinately, approaching the human--human reference of
0.659; the over-attribution disappears. The gap is therefore largely
attributable to the prompt, not the model.
Third, stratifying the one directional asymmetry led us to a defective
benchmark item: the inspector labels Collapse 1.81 times as often as native
speakers in our review sample, but only 1.17 times outside a single two-problem round, where
a load-bearing clause is deleted from the Bengali and Chinese translations,
leaving a question and a reference answer that are mutually incompatible. That
defect explains much but not all of the pooled asymmetry, and we report both
figures. We recommend four cheap checks before failure-mode rates are compared
across models: report label co-occurrence,
permute the tie-break, validate on a shared independent sample, and audit
systematic disagreements.
\end{abstract}


% =====================================================================
\section{Introduction}
\label{sec:intro}

A now-common move in LLM evaluation is to go beyond accuracy and characterise
\emph{how} models fail. A taxonomy of failure modes is defined, a judge
model applies it to transcripts, and the resulting distribution is reported
per model and compared across models. The appeal is obvious: accuracy alone
does not distinguish a model that misreads a problem from one that reasons
correctly and then slips.

This paper began as such a study. We defined a failure mode we call
\textbf{Compositional Collapse}: the model explicitly notices a required
structural sub-part of a problem (a rate that changes partway through, a
quantity that must be excluded) and then drops it or misapplies it
wholesale in a later step. It is distinct from never grasping the structure,
and distinct from a local arithmetic slip. We built a two-stage pipeline,
applied it to twelve models across eleven languages on a public benchmark,
and obtained the kind of result such papers report: Collapse appeared to be
the dominant failure mode everywhere we looked.

Verification did not support that result; this paper reports
what we found instead. Our contributions:

\begin{enumerate}
\item \textbf{A tie-break sensitivity result.} Categories in our taxonomy
co-occur on 50.1\,\% of failures. Because the classifier assigns one label per
item through ordered tests, the reported plurality is determined by test
order: swapping two tests moves the nominal plurality in eleven of twelve
models, and sweeping all six effective orderings moves Collapse's share of
failures across a fifty-point range (Section~\ref{sec:precedence}).

\item \textbf{A reliability comparison with a prompt-level diagnosis.} We
compare our inspector against a
second frontier model, an
open-weight model, and native speakers of all eleven languages
(179 transcripts), then run a \emph{second} independent native speaker per
language over the same 142 categorisable items. Both model judges
were run on the identical 179-item sample. The first human round and the
model raters agree closely with the inspector on answer correctness
($\kappa = 0.854$--$0.932$) but weakly on category assignment
($\kappa = 0.049$--$0.348$). On the 121 pairs with determinate second-round
labels, the two human rounds agree with \emph{each other} at
$\kappa = 0.659$. Re-running the same inspector model with the human
instructions verbatim raises agreement with the second human round from
$\kappa = 0.147$ to $\kappa = 0.636$ on the 106 of those pairs that it labels
determinately, closing most of the gap and eliminating the
over-attribution. The failure is therefore attributable to the instrument,
not the judge model
(Section~\ref{sec:reliability}).

\item \textbf{A benchmark defect found by chasing a directional
disagreement.} Pooled, the inspector labels Collapse 1.81 times as often as
native speakers. That excess is not general: outside one two-problem round the
factor is 1.17, and inside it 4.18. Investigating the difference revealed that
a load-bearing clause is deleted from two of that round's translations, so
models solve a different problem than the reference answer encodes and the
inspector scores the discrepancy as a reasoning failure. We report this as a
worked case of a benchmark defect masquerading as a model failure mode, and as
a caution that a directional judge--human gap is worth tracing to its items
instead of being reported as a rate (Sections~\ref{sec:reliability},
\ref{sec:translation}).

\end{enumerate}

We additionally document four benchmark-integrity defects in the
machine-translated source data (Section~\ref{sec:translation}).

We are not arguing that failure taxonomies are worthless. We are arguing that
the reliability of a category must be established before its rate is compared
across models, and that the checks required are cheap. We describe them so
they can be reused.

% =====================================================================
\section{Related Work}
\label{sec:related}

\textbf{Multilingual reasoning diagnostics.} Ranaldi and
Pucci~\cite{ranaldi2025} study multilingual mathematical reasoning and
cross-language performance; we use their benchmark and audit the reliability
of a downstream failure-taxonomy instrument applied to it. Ovalle et
al.~\cite{ovalle2025} check whether stated reasoning justifies answers on
multiple-choice items. Both treat the judgement of reasoning quality as
sufficiently reliable to report; our results examine that assumption for one
such instrument.

\textbf{Symbolic perturbation benchmarks.} GSM-Symbolic~\cite{mirzadeh2024}
established that templated perturbation of GSM8K~\cite{cobbe2021} items
exposes brittleness masked by aggregate accuracy. We use its multilingual
extension, and Section~\ref{sec:translation} reports defects introduced by
that extension's translation pass.

\textbf{LLM-as-judge reliability.} Zheng et al.~\cite{zheng2023} established
that strong models approximate human preference judgements well, and
subsequent work documents position, verbosity and self-preference
biases~\cite{panickssery2024,pombal2026}. That literature concerns
\emph{preference} between responses. Our setting is different and, we argue,
harder: assigning a causal-mechanistic label to a single transcript. Closest
to our position, Bavaresco et al.~\cite{bavaresco2025} evaluate eleven models
against human annotations on twenty tasks and find that judge reliability
varies substantially with \emph{the property being evaluated}, concluding that
LLM judges must be validated per task, not assumed. We extend that
result in two directions: we obtain non-model raters on the same items, which
reveals that the failure is directional, not merely noisy
(Section~\ref{sec:reliability}), and we identify a specific mechanism,
tie-break order under category co-occurrence, that generates the
instability (Section~\ref{sec:precedence}).

\textbf{Reasoning-error taxonomies.} Typologies of reasoning errors applied by
automated raters are established practice~\cite{golovneva2023roscoe}, and
LLM-based scoring of intermediate steps instead of final answers is
already in use~\cite{xia2025reasoneval}. Step-level annotation schemes
routinely assign \emph{multiple simultaneous labels} to a single step by
design~\cite{jacovi2024reveal}, and a recent survey reports that evaluation
criteria in this area overlap and that practice is inconsistent across
studies~\cite{lee2025survey}. Separately, Tyen et al.~\cite{tyen2024} find
that LLMs struggle to locate reasoning errors even in objective, unambiguous
cases. Our results are consistent with both observations and quantify their
cost for one taxonomy.

\textbf{Faithfulness of stated reasoning.} Compositional Collapse is a
specific case of a broader phenomenon: stated reasoning that does not govern
the computation which follows~\cite{turpin2023,chen2025reasoning}. That
literature asks whether a verbalised step reflects the underlying
computation; we ask the prior question of whether independent raters can agree
on which transcripts exhibit such a step at all.

\textbf{Adjacent uses of ``collapse''.} We distinguish two nearby uses.
\emph{Composition Collapse}~\cite{yu2026composition} is a failure to chain two
stable facts, not the abandonment of a step already taken.
\emph{Cross-lingual Collapse}~\cite{park2025crosslingual} is chain-of-thought
reverting to the dominant pre-training language. Ours is narrower: a required
sub-part is represented in the trace and then dropped or misapplied. The
naming overlap is itself a reason to report reliability with the construct.

% =====================================================================
\section{Method}
\label{sec:method}

\subsection{Data}

We use mGSM-Symbolic (the multilingual symbolic variant of
GSM8K~\cite{cobbe2021} in the line of MGSM~\cite{shi2023mgsm}), released
alongside Ranaldi and Pucci~\cite{ranaldi2025} as part of the SWATH project.
The dataset is available under CC-BY-SA 4.0 on HuggingFace
(\texttt{lrana/MGSM-Symbolic}). The translation defects documented in
Section~\ref{sec:translation} have been reported to the maintainers.
Our battery
comprises 22 problem instances drawn from \textbf{eight
distinct structures}: Round~1 is a single fog-bank template under eight
numeric substitutions; Round~2 is four templates with two instances each;
Round~3a is two templates with two each; Round~3b is one template with two,
carrying an additional strict output-format requirement. We state the
structure count, not the instance count, because the smaller number is
the honest description of the design's breadth.

Rounds 1, 2 and 3a share a nine-language set (Chinese, Thai, Japanese, German,
Spanish, French, Italian, Russian, Swahili); Round~3b substitutes Bengali and
Telugu for Japanese and German. Across the battery this is eleven language
codes and up to 198 language-instances per model.

\subsection{Pipeline}

\textbf{Stage 1 (blind generation).} The model under test solves problems with
a generic math-assistant system prompt, no failure-mode coaching, and no
prompt engineering. Each problem is issued as an isolated call with tool
access disabled, one call per problem, so no in-context transfer across the
near-isomorphic instances within a round is possible. The output token budget is
64{,}000 for every model reported.

\textbf{Stage 2 (fixed inspection).} A single fixed model (Claude Sonnet~5,
never varied) receives the question, the reference answer, and the raw
output, and returns structured JSON: the extracted answer, a correctness
judgement, and a quoted-evidence field for each failure mode or the literal
string \texttt{none}. The full instrument is in Appendix~\ref{app:inspector}.
Only the generator varies across conditions.

\subsection{Taxonomy and Its Tie-Break}

\textbf{Compositional Collapse}: the model explicitly mentions a required
clause, then drops or misapplies it. \textbf{Reasoning Drift}: early steps
correct, a later step introduces a cascading error. \textbf{Format
Instability}: output structure breaks despite explicit requirements (scoped to
Round~3b by construction; the tag requirement was enabled for seven of twelve
models there, so its pooled rate has a mixed denominator). \textbf{Mismodeling}:
residual.

The evidence fields are populated \emph{independently}, so one transcript can
carry evidence for several modes. This is not unusual: step-level reasoning
annotation schemes assign several labels to a single step by
design~\cite{jacovi2024reveal}. To produce a single label per item the
classifier tests categories in a fixed order and returns on the first match,
with Collapse tested first. Section~\ref{sec:precedence} is about the
consequences of that choice, which we did not initially treat as consequential.

\subsection{Model Panel and Its Confounds}

The twelve models span three proprietary API systems and nine locally-served
open-weight models. This panel is a convenience sample and is confounded in
ways that matter for interpretation, so we state them instead of letting a
capability-range reading stand; unstated sampling of this kind is a
documented source of invalid benchmark claims~\cite{bean2025}. The API and local groups are perfectly
separated by accuracy: the lowest API model (67.18\,\%) exceeds the highest
local model (54.04\,\%) with no overlap. Within the local group, parameter
count does \emph{not} predict accuracy ($r = 0.025$, $p = 0.95$); a 14B
reasoning-tuned model reaches 54.04\,\% while a 34B general model reaches
6.91\,\%. Reasoning post-training tracks accuracy far better than scale does
(local reasoning-tuned mean 43.0\,\% versus 10.1\,\% otherwise). Local models
were served through a runtime whose default quantisation we did not record,
so their precision is unknown. Deployment class, scale, reasoning training
and quantisation therefore vary together and none is controlled. No claim in
this paper depends on the panel forming a capability axis. We also note that
the fixed inspector (Claude Sonnet~5) is itself a panel model, and a second
panel model (Claude Haiku~4.5) shares its vendor family; the implications for
self-inspection and same-family bias are discussed in
Section~\ref{sec:limitations}.

% =====================================================================
\section{Results}
\label{sec:results}

\subsection{The Construct as Measured}
\label{sec:descriptive}

Twelve models at up to 198 language-instances each give 2{,}376 attempted
records, of which we analyse \textbf{2{,}316}. The 60 excluded are pipeline
failures, not model behaviour: 43 are inspector calls that returned an
unfilled JSON schema instead of a verdict, and 17 carry an explicit
generation- or inspection-error marker. They are unevenly distributed
(Qwen2.5-32b loses 28, Neural-Chat-7b 13, Yi-34b 10, and five models lose
none), so per-model denominators differ and are given in
Table~\ref{tab:precedence}. Note also that 2{,}316 records are not 2{,}316
independent problems: they instantiate eight structures under numeric
substitution across eleven languages.

Accuracy ranges from 0.54\,\% (Neural-Chat-7b) to 70.20\,\% (Claude
Haiku~4.5). Under our classifier, Compositional Collapse is the most frequent
single failure category in every model. Read naively, this is a strong result about models.
The remainder of this section is about why it is mostly a result about the
classifier.

\subsection{The Plurality Is Set by the Tie-Break}
\label{sec:precedence}

Because evidence fields are populated independently, we can ask how often
they co-occur. Across the twelve models, \textbf{50.1\,\% of failures}
(775 of 1{,}546) carry evidence for more than one category, and the overlap is
almost entirely one pair: 761 of the 775 are exactly Collapse together with Drift.

Re-running the identical classification with Drift tested first, changing
nothing else, moves the nominal plurality from Collapse to Drift in
\textbf{eleven of twelve models} (Table~\ref{tab:precedence}). The statement
``Collapse is the dominant failure mode'' is therefore close to a restatement
of ``we tested Collapse first.''

\begin{table*}[t]
\centering
\caption{The same classifier under two orderings of its tests. $N$ is analysed
records; Plur.\ is the plurality category and Share its percentage of that
model's failures; Dual is the share of failures carrying evidence for more
than one category. Nothing changes but the order in which two tests are
applied.}
\label{tab:precedence}
\footnotesize
\begin{tabular}{lrrlrlrr}
\toprule
& & & \multicolumn{2}{c}{Collapse-first} & \multicolumn{2}{c}{Drift-first} & \\
\cmidrule(lr){4-5}\cmidrule(lr){6-7}
Model & $N$ & Fails & Plur.\ & Share & Plur.\ & Share & Dual \\
\midrule
Claude Haiku 4.5 & 198 & 59 & Collapse & 66.1 & Drift & 54.2 & 30.5 \\
DeepSeek-V4-Pro & 198 & 62 & Collapse & 50.0 & Drift & 77.4 & 38.7 \\
Claude Sonnet 5 & 195 & 64 & Collapse & 62.5 & Collapse & 51.6 & 10.9 \\
Phi4-reasoning:14b & 198 & 91 & Collapse & 63.7 & Drift & 56.0 & 35.2 \\
Deepseek-r1:32b & 197 & 117 & Collapse & 64.1 & Drift & 53.0 & 30.8 \\
Qwen2.5-32b & 170 & 108 & Collapse & 75.0 & Drift & 55.6 & 38.0 \\
Deepseek-r1:14b & 198 & 130 & Collapse & 72.3 & Drift & 56.2 & 41.5 \\
Mistral-Nemo-12b & 196 & 175 & Collapse & 78.3 & Drift & 74.3 & 57.1 \\
Yi-34b & 188 & 175 & Collapse & 89.1 & Drift & 65.7 & 58.3 \\
phi3.5-latest & 195 & 187 & Collapse & 81.3 & Drift & 72.7 & 62.0 \\
Mixtral-8x7b & 198 & 194 & Collapse & 78.4 & Drift & 69.6 & 57.7 \\
Neural-Chat-7b & 185 & 184 & Collapse & 90.2 & Drift & 73.9 & 72.3 \\
\bottomrule
\end{tabular}
\end{table*}

\textbf{All six effective orderings.} Two orderings show the plurality is
fragile; sweeping every ordering shows how much of the reported distribution
is a free parameter. Three of our four categories carry an independent
evidence test; Mismodeling is a residual, matched only by items with no
evidence for anything, so its position in the sequence cannot change any
assignment. The $4! = 24$ permutations therefore reduce to $3! = 6$ distinct
classifiers, and we report those six.

Across them, Compositional Collapse takes any value between \textbf{26.4\,\%}
and \textbf{76.4\,\%} of failures, and Reasoning Drift between 14.9\,\% and
64.7\,\%, swings of roughly fifty percentage points each. The plurality
winner is Collapse under three of the six orderings and Drift under the other
three. The residual category is invariant at 8.5\,\% for the reason just
given; Format Instability is scoped to one round by design and moves only
between 0.0\,\% and 0.9\,\%. A reader given one number from this range has no
way to know which of the six they were shown.

We therefore withdraw the ordered claim and report the ordering-invariant
quantity instead: the share of failures carrying Collapse evidence at all.
That ranges from 50.0\,\% to 90.2\,\% and exceeds half of failures in eleven
of twelve models. This quantity is still the inspector's rate ---
Section~\ref{sec:reliability} shows that the inspector's per-item Collapse
assignments are not reproduced by independent raters --- and we report it as
descriptive of this instrument's output, not as a supported claim about model
behaviour.

Two observations extend beyond our taxonomy. Category ambiguity is not
confined to marginal cases: at 50.1\,\% it is the modal situation, which is
consistent with schemes that assign several labels per step by
design~\cite{jacovi2024reveal} and with survey evidence that criteria in this
area overlap~\cite{lee2025survey}. And any
evaluation assigning one failure label per item via ordered tests is exposed
to the same effect and will not detect it unless the order is deliberately
permuted, a check costing one additional pass over existing verdicts.

Finally, ambiguity is not uniform: the dual-evidence rate spans 10.9\,\%
(Claude Sonnet~5) to 72.3\,\% (Neural-Chat-7b), and the higher rates fall on
the lower-accuracy models. Because four of the twelve models are near-random
on this benchmark (accuracy below 10\,\%) and contribute 740 of the 1{,}546
failures, we also report the restricted figures: on the eight models at or
above 10\,\% accuracy the dual-evidence rate is 38.7\,\% (312/806, against
62.6\,\% below the threshold) and the share of failures carrying Collapse
evidence is 68.9\,\% (against 84.6\,\% below). The co-occurrence phenomenon is
therefore not carried by the weakest models, though they amplify it; the
agreement result also holds on these models (Table~\ref{tab:robust}). We report
the accuracy--ambiguity association as an observation, not a
finding: twelve confounded model-level points cannot support a mechanistic
reading. But it bears on interpretation, because a single-label
distribution means least where evidence overlaps most.

\subsection{Raters Agree on the Answer, Not the Reason}
\label{sec:reliability}

The preceding result concerns one classifier's internal logic. The sharper
question is whether independent raters identify
the same instances. We report Cohen's $\kappa$~\cite{cohen1960} throughout. We compared our fixed
inspector against three further rater sets: a second frontier model
(DeepSeek-V4-Pro) and an open-weight model (Qwen2.5-32b) --- both themselves
members of the panel under test --- and
\textbf{native speakers of all eleven languages} who judged 179 traces, one
reviewer per language.

Traces were drawn by a seeded script (seed 42) that stratifies across
languages, models and inspector outcomes (correct versus incorrect, and
with versus without each failure-mode evidence field), so the frame is not
independent of inspector labels: it is balanced on them by construction, which
means the sample over-represents flagged items relative to the corpus and the
marginal rates below should not be read as corpus rates. Traces whose raw
output exceeded a character cap were excluded to bound reviewer reading time;
this falls hardest on the most verbose model (phi4-reasoning:14b), which
remains in the sample at reduced representation (14 of 179 items). Bengali
and Telugu, given one item each by that draw, were instead reviewed in full
(31 and 30 traces; 61 of the 179), with eight long outputs cut to
4{,}000 characters rather than excluded. Reviewer documents were then issued per language. We record one
provenance caveat: the number of items per language was not equalised
(categorisable items: Bengali 31, Russian 8). The model judges originally graded a different
pinned sample; both were subsequently re-run on this identical
179-item set, so all comparisons in Tables~\ref{tab:reliability}
and~\ref{tab:humanhuman} are crossed on the same records.

Human reviewers and the automated inspector received the same three inputs
(apart from those eight shortened outputs): the translated question, the
numeric reference answer, and the model's raw output. Reviewers were asked two questions in plain language (whether the
final answer was correct, and, if not, whether the model had noticed a key
detail and then dropped or misused it), together with a plain-language
operational definition of that pattern and one illustrative example
constructed for the instructions. They were shown neither the category name
nor the inspector's verdict. This symmetry matters for
the Round~3b analysis below: where the two procedures diverge, it is not
because one held information the other lacked.

This symmetry is only partial, however, and the difference turns out to
matter: the two procedures share inputs and a core definition, but differ in
definition wording, framing and response format, example, and anchoring to the
reference answer (Section~\ref{sec:limitations} lists the four asymmetries). The disagreement
in Table~\ref{tab:reliability} is therefore not attributable to the human
lacking information --- but neither, as the prompt-level experiment below
shows, is it independent of how the task is presented. One alternative
explanation for the \emph{direction} of the gap, which we cannot rule out,
concerns the illustrative
example: it is a clean canonical case (a quantity computed and then omitted
from the final step), and reviewers anchored on it may have applied a
narrower reading than the inspector, which would depress the human rate
without either rater being wrong. We report this as a live confound rather
than a resolved one.

\begin{table}[t]
\centering
\caption{Agreement with the fixed inspector. Raters are native speakers of
all eleven languages, a frontier judge model, and an
open-weight judge model. All comparisons show high agreement on answer
correctness and substantially lower agreement on whether the failure is
Compositional Collapse. All raters graded the identical 179-item human
sample (correctness $\kappa$ on 178 joined items; Collapse $\kappa$ on the
142 items the human rater judged incorrect).}
\label{tab:reliability}
\footnotesize
\begin{tabular}{lrrr}
\toprule
Rater & $n$ & Answer $\kappa$ & Collapse $\kappa$ \\
\midrule
Native speakers & 179 & \textbf{0.932} & \textbf{0.093} \\
DeepSeek-V4-Pro & 179 & 0.917 & 0.348 \\
Qwen2.5-32b     & 179 & 0.854 & 0.049 \\
\bottomrule
\end{tabular}
\end{table}

% Figure removed for camera-ready: it duplicated Table~\ref{tab:reliability}
% (reviewer comment), and the expanded tables now carry the content.
% The figure file remains at figures/fig_reliability.pdf if reinstated.

Table~\ref{tab:reliability} is the paper's central result. On answer
correctness, agreement is near-perfect for all three rater sets, including exact
agreement with humans on 97.8\,\% of 178 joined items. This establishes that the pipeline
works, that the raters are competent, and that the transcripts are legible.
The four disagreements are themselves informative, and all four trace to
truncated text rather than to raters reading the same words differently. In
two, the inspector scored the answer correct where the human reported that no
final answer existed --- but the review copy of each transcript had been
shortened by our own document preparation, which inserted a truncation marker
at about 4{,}000 characters, so the reviewer never saw the final answer the
inspector found in the full text. In the other two the direction reverses,
the human scoring an answer correct where the inspector recorded none: one
transcript carries the max-tokens sentinel, the other ends mid-sentence at
28{,}176 characters with no marker at all. Truncation therefore corrupts the
comparison in both directions --- hiding answers that were reached, and ending
generations before any answer --- and it need not leave a sentinel to audit.
This
is a stronger reason to regenerate affected runs than to discard them.
On category assignment, agreement falls to $\kappa = 0.049$--$0.348$ for all
three.

The human row constrains what the low agreement can be attributed to. Native
speakers agree with the inspector on answer correctness at $\kappa = 0.932$
(97.8\,\% of 178 joined items) and on category assignment at $\kappa = 0.093$
(50.7\,\% of 142 items). Eligibility for the second question was set by the
\emph{human} rater: reviewers were asked about noticing-and-dropping only when
they had judged the answer incorrect, which they did on 142 of 179 items. The
category comparison is therefore conditioned on the human's correctness
judgement, not the inspector's. The category $\kappa$ of 0.093
is near chance for a binary judgement. We give the numeric coefficients,
not the Landis--Koch verbal bands~\cite{landis1977}, which are a
convention rather than a calibrated scale~\cite{artstein2008}. Stated
precisely: the human labels did not reproduce the inspector's item-level
Collapse assignments. We have no independent ground-truth rater, so this
establishes non-reproducibility, not that either rater is correct.

This is not an artifact of chance correction. Cohen's $\kappa$ is known to
behave poorly when marginals are skewed, and ours are: the inspector labels
Collapse almost twice as often as the human. But the well-known paradox
involves high raw agreement paired with low $\kappa$, whereas raw agreement
here is 50.7\,\%, against 45.7\,\% expected by chance given the skewed
marginals. Two statistics
designed to be robust to marginal skew agree: PABAK~\cite{byrt1993}
$= 0.014$ and Gwet's AC1 $= 0.015$. The $2 \times 2$ table is 38 items both raters call Collapse,
34 neither, 56 inspector-only and 14 human-only.

\textbf{A second independent human round separates the two explanations.} Low
inspector--human agreement is consistent with two very different diagnoses:
the category may be intrinsically ambiguous, in which case no rater could
apply it consistently, or the inspector may simply be an outlier. These are
distinguished by a second human rater on the same items, which we ran. A
different native speaker per language independently re-rated all 142
categorisable items, blind to the first reviewer's labels, to the inspector's
verdicts and to the category name, using the same Collapse definition and
response options. As noted in Section~\ref{sec:limitations}, document handling
differed between rounds, so we do not claim the instruments were strictly
identical. The two rounds also differ procedurally: the first round asked the
category question only where that reviewer had judged the answer incorrect,
whereas the second re-rated all 142 items preselected on that basis.

\begin{table}[t]
\centering
\caption{Agreement on the Compositional Collapse label among the 121 items
with determinate labels from both human rounds. The second reviewer marked 21
of the original 142 items ``Unclear''; these are excluded here and analysed
separately in the text. Human~1 and Human~2 are the review Round~1 and
Round~2 reviewers. The inspector and both judge models graded the identical
179-item sample; the H-style inspector re-graded the 142 categorisable items. ``H-style insp.'' is the same model as the fixed inspector
(Claude Sonnet~5) but using the human instructions verbatim; all other
automated raters use the original inspector prompt.}
\label{tab:humanhuman}
\footnotesize
\begin{tabular}{lrrr}
\toprule
Rater pair & $n$ & $\kappa$ & Raw agr. \\
\midrule
\textbf{Human 1 vs.\ Human 2} & 121 & \textbf{0.659} & 83.5\,\% \\
\midrule
Human 1 vs.\ inspector       & 121 & 0.074 & 50.4\,\% \\
Human 2 vs.\ inspector       & 121 & 0.094 & 52.1\,\% \\
DeepSeek vs.\ Human 1         & 121 & 0.064 & 51.2\,\% \\
DeepSeek vs.\ Human 2         & 121 & 0.090 & 52.9\,\% \\
DeepSeek vs.\ inspector       & 121 & 0.333 & 69.4\,\% \\
Qwen vs.\ Human 1             & 121 & $-0.134$ & 51.2\,\% \\
Qwen vs.\ Human 2             & 121 & $-0.065$ & 52.9\,\% \\
Qwen vs.\ inspector           & 121 & 0.053 & 39.7\,\% \\
\midrule
\textbf{H-style insp.\ vs.\ Human 1} & 106 & \textbf{0.537} & 77.4\,\% \\
\textbf{H-style insp.\ vs.\ Human 2} & 106 & \textbf{0.636} & 82.1\,\% \\
H-style insp.\ vs.\ inspector  & 106 & 0.197 & 58.5\,\% \\
\bottomrule
\end{tabular}
\end{table}

The separation is clear (Table~\ref{tab:humanhuman}). The two independent
humans agree at $\kappa = 0.659$ (83.5\,\% raw; bootstrap 95\,\% CI
$[0.514, 0.785]$, and $[0.496, 0.785]$ under a language-clustered bootstrap),
while each agrees with the original inspector at $\kappa \leq 0.094$.
Running the two model judges on the identical 179-item sample confirms
that this is not a sample artifact: DeepSeek-V4-Pro, cross-matched to
the human set, agrees with the humans at $\kappa = 0.064$--$0.090$, closely
matching the original inspector's $0.074$--$0.094$, and Qwen2.5-32b
disagrees in the opposite direction, \emph{under}-attributing Collapse
(13.2\,\% against the humans' 40.5--42.1\,\%) with $\kappa < 0$. The
original-prompt judges thus scatter around the human rate in both
directions. Human--human
disagreements are nearly symmetric (9 items where only Human~1 says
Collapse against 11 where only Human~2 does), so neither human is the more
liberal rater. The separation holds under every subset we examined: excluding
Round~3b, human--human is 0.649 against 0.171 for human--inspector; within
Round~3b alone, 0.641 against 0.051; excluding Bengali and Telugu, 0.646
against 0.172.

We note that both human rounds received the same illustrative example, and
two raters anchored on it may agree because they applied the same narrow
reading. The result should therefore be read as ``reproducible under this
definition and example'' rather than as a property of the construct
independent of its instructions. A round with a different example, or none,
would test this.

Twenty-one of the 142 items drew an explicit ``Unclear'' from the second
reviewer and are dropped above. The conclusion does not depend on that choice:
recoding all Unclear responses as \emph{No} gives $\kappa = 0.650$, while
recoding all of them as \emph{Yes} gives $0.467$, still above the reported
human--inspector values. We note that review Round~1 offered the equivalent option
(``Not sure'') and no
reviewer used it in 179 judgements, whereas review Round~2 used it 21 times; we
cannot say whether that reflects the reviewers or the instructions.

\textbf{This changes what the paper can claim.} The construct is not
unmeasurable: two independent native speakers, given the same short
operational definition and example and no other coordination, reproduce each other's
Collapse judgements at $\kappa = 0.659$. What fails to reproduce is the
\emph{automated inspector's} application of it.

\textbf{Diagnosing the instrument: a prompt-level experiment.} The preceding
comparisons show \emph{that} the inspector disagrees with humans; they do
not distinguish whether the cause is the judge model, the response format,
or the prompt. Four asymmetries between the procedures are identifiable in
retrospect (Section~\ref{sec:limitations}), so we ran the obvious control:
re-grade the 142 categorisable items with \emph{the identical model}
(Claude Sonnet~5) but using the exact human instructions --- same narrower
definition (\emph{noticed} only, without ``or invented''), same illustrative
example (the apples item), the same binary yes/no question conditioned on a
prior judgement of incorrectness. Nothing else changes: the same 142 items,
the same three inputs (question, reference answer, raw output).

The result is shown in Table~\ref{tab:humanhuman}. The human-style inspector
answered ``Unclear'' on 15 of the 121 human-determinate pairs (26 of 142
overall --- it uses the option the original inspector lacked), leaving 106
pairs on which both it and the humans gave determinate labels. On those 106
pairs it agrees with Human~2 at
$\kappa = 0.636$ (82.1\,\% raw; the original inspector reaches 0.147 on the
same pairs), approaching the human--human reference
($\kappa = 0.659$, 83.5\,\%). The collapse rate matches the humans' as well
(44.3\,\% against 40.5--42.1\,\%), with no directional over-attribution.
Yet this inspector agrees with the \emph{original} inspector at only
$\kappa = 0.197$, despite being the same model. The same model, given
the human instructions instead of the original prompt, reproduces the human
labels. The same model, given the original prompt, does not. The finding is
therefore about the instrument, not the judge: the gap is largely
attributable to the four prompt asymmetries identified in
Section~\ref{sec:limitations}, and eliminating them eliminates most of it.

\textbf{The disagreement is directional, and its direction depends on the
text being judged.} Pooled over all items, the inspector labels
\textbf{66.2\,\%} of failures Compositional Collapse where native speakers
label \textbf{36.6\,\%}, a factor of 1.81, and item-level disagreements
run \textbf{four to one} in one direction: 56 items where the inspector claims
Collapse and the human does not, against 14 the other way.

That 1.81 is a property of this sample, not an estimate for the benchmark.
Because our per-language sample sizes are unequal (Bengali alone
contributes 31 of 142 categorisable items), the pooled figure is weighted
toward the languages where disagreement is largest. Weighting each language
equally instead gives \textbf{1.38}. We report the sample figure as a sample
figure throughout, and make no corpus-level claim about the size of the
inflation.

Splitting the comparison by language is what isolated the problem. Three languages (Telugu, Swahili and Bengali) account for essentially all of the directional gap, the inspector exceeding the human by 57--71 points in each, while the remaining eight average to no net offset. We treat this split as diagnostic, not as an estimate of a language effect: with one reviewer per language, reviewer and language are perfectly confounded (Section~\ref{sec:limitations}). Its value was that it told us which items to read.

\textbf{That pattern should not be read as a gradient, and chasing it
found a benchmark defect.} Bengali and Telugu appear only in Round~3b, which
contributes 57 of the 142 categorisable items (40\,\% of the human
comparison) from just two distinct problems. Table~\ref{tab:robust} splits
the comparison on that round. Outside it the inspector and the native speakers
almost agree on marginal rates (56.5\,\% against 48.2\,\%, a factor of 1.17);
inside it the factor is 4.18. The eight languages outside the three lowest
show no internal gradient at all ($r = -0.527$, $p = 0.18$), so the pooled
association describes three separated points, not a slope.

\begin{table}[t]
\centering
\caption{Human--inspector comparison under successive exclusions. Agreement on
the category stays far below correctness agreement (Table~\ref{tab:reliability}) under every restriction, and among the exclusion rows is lowest on the
core models alone; the over-attribution factor does not survive. Telugu (21 of 25
intact Round~3b items) was reviewed by the author; see
Section~\ref{sec:limitations}.}
\label{tab:robust}
\footnotesize
\begin{tabular}{lrrrr}
\toprule
Subset & $n$ & $\kappa$ & Insp.\ & Hum.\ \\
\midrule
All categorisable items & 142 & 0.093 & 66.2 & 36.6 \\
Excluding Bengali & 111 & 0.161 & 61.3 & 43.2 \\
\textbf{Excluding Round 3b} & \textbf{85} & \textbf{0.180} & \textbf{56.5} & \textbf{48.2} \\
Core models only & 112 & 0.057 & 69.6 & 37.5 \\
Core models, $\geq$10\,\% accuracy & 66 & 0.066 & 71.2 & 37.9 \\
\textbf{Core models, excl.\ Round 3b} & \textbf{67} & \textbf{0.115} & \textbf{58.2} & \textbf{46.3} \\
\midrule
Round 3b only & 57 & 0.108 & 80.7 & 19.3 \\
\quad --- of which bn, zh & 32 & 0.052 & 84.4 & 12.5 \\
\quad --- other languages & 25 & 0.219 & 76.0 & 28.0 \\
Excluded-model items only & 30 & 0.217 & 53.3 & 33.3 \\
\bottomrule
\end{tabular}
\end{table}

\textbf{The result is not an artifact of the excluded runs.} Twenty-three per
cent of the 179 review Round~1 transcripts (41 items) --- or 21\,\% of the 142
categorisable items (30 items) --- come from the four models withheld from
the core table for truncation (Section~\ref{sec:limitations}), which raises the
obvious concern that low agreement reflects unreadable output, not an
unreliable category. It does not, and the direction is the opposite of the
concern: on those items agreement is the \emph{highest} of any subset
($\kappa = 0.217$), and removing them lowers agreement on the remainder to
$\kappa = 0.057$. A plausible reading is that when a transcript is visibly
empty both raters agree easily that nothing was noticed and dropped. We note
that ``core models'' is a claim about the generation harness, not about
transcript quality: core-model output can still be malformed, and can still
come from a defective benchmark item, as Round~3b shows.

Inspecting Round~3b's two items accounts for much of the remaining gap,
though not all of it. The
English specifies that Ibrahim's dogs each have ``13 puppies \emph{with
spots}''; that clause is what places all 702 of his puppies in the numerator.
It survives translation into Italian, French, Russian, Spanish, Swahili, Thai
and
Telugu, and is \emph{absent} from the Bengali and Chinese versions of both
items. Numeric tokens are identical across every language version of these
items, so the check
in Section~\ref{sec:translation} does not detect it. Bengali scores
\textbf{0.0\,\%} on this round across all twelve models, including the three
frontier systems that exceed 67\,\% overall, as does Chinese, which reaches
30.6\,\% elsewhere. Bengali output length is normal (median 1{,}037
characters) and no transcript carries a truncation sentinel, though four of
the 46 Bengali and Chinese transcripts on this round stop mid-sentence (three
from phi3.5-latest, one from yi-34b; all four are scored incorrect). The rest
reason at length and reach the answer the corrupted text supports.

The arithmetic is checkable without reading Bengali: the reference answer is
$801/900 = 89\,\%$, which requires Ibrahim's 702 puppies to count as spotted,
whereas the corrupted text supports $99/900 = 11\,\%$. Eleven per cent is what
the Bengali transcripts report, and it is what our reviewers saw.

Both rating procedures received the same three inputs on these items (the
translated question, the numeric reference answer, and the model response,
though four Bengali review copies were shortened), so the divergence is not explained by one rater holding information the other
lacked. What the defect creates is a question and a reference answer that are
mutually incompatible, and under that condition the two procedures applied the
noticed-and-dropped criterion differently. The inspector's own quoted evidence
on these items is consistent with the English semantics, not with the
Bengali text: on one it faults the model for treating ``Ibrahim's puppy-count
data as if it were spot-count'', which is the correct reading of the
Bengali as written, since the attribute is absent there. Stratified
disagreement identified the item for audit; direct inspection of the source
and its translations established the defect. The 6.50 ratio on Bengali
(83.9\,\% inspector against 12.9\,\% human) is the
signature of a defective item, not of a low-resource language.

The second human round, run after this analysis and blind to it, corroborates
the reading independently. On the Chinese instance of the same item the
inspector records Collapse; both human rounds record no Collapse; and the
second reviewer's unprompted note reconstructs the defect exactly: the
model ``counted Ibrahim's 702 puppies in the total but treated them as 0
spotted because spots were not mentioned.'' Across the 32 Bengali and Chinese
instances of these two items, the inspector assigns Collapse in 27 while the
two human rounds assign it in 4 and 8 respectively. Whatever the inspector is
detecting on this item, two independent native speakers do not see it, and one
of them names the translation as the reason without being asked about it.

The two corrupted translations do not account for the whole of the Round-3b
asymmetry. Within that round the intact languages still show 76.0\,\% against
28.0\,\% ($\kappa = 0.219$), so substantial disagreement remains after the
defect is set aside, though 21 of those 25 items are Telugu, and Telugu is the
language reviewed by the author (Section~\ref{sec:limitations}), so we do not
read the residual as an independent effect either.

What survives is the reliability finding, which is what the paper rests on.
Agreement on the category rises from $\kappa = 0.093$ to $0.180$ when Round~3b
is removed and falls to $0.115$ on core models within that restriction; every
value remains far below the correctness coefficients. What does not survive is the inflation factor. A rate
1.81 times the human rate would be a serious bias in any reported taxonomy; a
factor of 1.17 on clean items is not. We report both, not the pooled
number alone, and we note the general lesson: a directional judge--human gap is
worth tracing to the items that produce it, because it may be measuring the
benchmark rather than the judge.

Whether a second model rater would have surfaced this item can be
answered directly (Table~\ref{tab:reliability}). DeepSeek-V4-Pro largely
sides with the inspector ($\kappa = 0.348$ against it on the categorisable
items, against $\kappa \leq 0.090$ against either human), while Qwen2.5-32b
sides with neither --- its Collapse rate of 13.2\,\% is below both the
inspector's and the humans'. Neither judge reproduces the human labels.

\subsection{Benchmark Translation Defects}
\label{sec:translation}

That translated evaluation data carries artifacts which confound cross-lingual
comparison is established~\cite{artetxe2020}, and recent work attributes
measurable accuracy drops specifically to target-side translation errors after
controlling for source-side correctness~\cite{thellmann2026}. Our check is a
lightweight detector for the same class of defect, requiring no annotation:
while constructing a monolingual control we compared the multiset of numeric
tokens across all nine translations of each item. The translations disagree
with each other. Four deviations are substantive:

\begin{itemize}
\item \textbf{Chinese, apartment-vacancy item.} The source specifies eight
apartments vacant in each of five complexes. The Chinese states that
\emph{every} apartment was vacant, deleting the quantity and changing the
answer from 40 vacant units to 500. A model reading only the Chinese cannot
reach the reference answer. Our Chinese reviewer, who saw the question, the
reference answer and the transcript but not the inspector's verdict,
independently judged the model wrong but \emph{not} an instance of
Compositional Collapse: the model had solved the problem it was given. The
defect is in the benchmark, and a failure-mode taxonomy applied to that item
measures the translation, not the model.
\item \textbf{Thai, both apartment-complex items.} The clause ``than the other
$N$ combined'' is dropped, leaving a comparison without a referent.
\item \textbf{Chinese, Round~1.} The rate clause is duplicated, adding a
numeral without adding information.
\item \textbf{Bengali and Chinese, both Round~3b items.} The English
attributes spots to Ibrahim's puppies (``13 puppies \emph{with spots}''),
which is what places all 702 of them in the numerator. Both translations
render the clause as ``13 puppies,'' deleting the attribute. Italian, French,
Russian, Spanish, Swahili, Thai and Telugu preserve it. Both affected languages score
0.0\,\% on this round across all twelve models; Chinese reaches 30.6\,\% in
its other rounds, and Bengali appears in no other
round. This defect was found by reading the items after the human
comparison flagged an anomaly (Section~\ref{sec:reliability}), \emph{not} by
the numeric-token detector.
\end{itemize}

Two further deviation classes are detector artifacts, not data defects, and we
note them so the method can be reused without overcounting: Bengali writes
digits in Bengali numerals, and Japanese renders fractions in a form that
emits as separate numerals what English writes as one token.

\textbf{A limit of the method.} Numeric-token
comparison detects deletions and duplications of quantities. It cannot detect
the deletion of an \emph{attribute} attached to a quantity, because the
quantity survives: the Bengali and Chinese Round-3b items carry numeric
multisets identical to the English. We attempted two language-agnostic
extensions (per-sentence length-share profiles, and counts of the most
repeated content token), and neither is usable. Both are dominated by
orthographic and morphological variation: Thai marks no sentence boundaries,
Swahili segments differently, and inflected languages fragment repeated words,
so both detectors flag high-resource languages with intact translations as
readily as damaged ones. We therefore report no automated method for this
defect class, and note that the number of such defects remaining in this
benchmark is unknown.

Three of the four substantive defects fall in languages sitting mid-table by
accuracy; the fourth is the direct cause of the only two 0.0\,\% language--round
accuracy cells (Bengali and Chinese on Round~3b). Pooled across the twelve models, accuracy ranges from
53.12\,\% in Italian to 16.29\,\% in Swahili across the nine languages of
Rounds~1, 2 and~3a. But with defective items now known in three of the eleven languages
(Chinese, Thai and Bengali) --- two of them inside the nine-language ordering
above --- that ordering should not be read as a language-capability ranking
without item-level verification. The general
implication stands: a cross-lingual
accuracy gap measured on machine-translated data may be measuring translation,
a failure-mode taxonomy applied to a defective item measures translation as
well, and the cheap numeric check catches only part of the problem.

% =====================================================================
\section{Discussion}
\label{sec:discussion}

Our results separate two things that a single agreement coefficient conflates.
Judgements that are \emph{verifiable} (did the model produce the right
number?) were highly reliable across raters of very different kinds, from a
32B open-weight model to native speakers of eleven languages
($\kappa$ 0.854--0.932). Judgements that are \emph{interpretive} (did the
model notice something and then abandon it?) were not reproduced by the
inspector under its original prompt, in any comparison we ran.

The natural inference from that alone would be that interpretive judgements
are intrinsically unreliable. Our second human round shows that inference is
wrong, at least here. Two native speakers who never saw each other's labels,
working from a short operational definition, agreed on the interpretive
judgement at $\kappa = 0.659$. The category can be applied consistently by
people under this definition and example; what does not reproduce is one automated instrument's application of
it. The distinction matters for practice: a low judge--human $\kappa$ may be
evidence about the judge or its prompt rather than about the category, and
it takes a second human to tell the two apart.

This matters because the LLM-as-judge literature establishing high agreement
concerns preference between responses~\cite{zheng2023}, a comparative
judgement. Assigning a causal-mechanistic label to a single transcript is a
different task, and our results indicate its reliability must be established
separately, not inherited. Bavaresco et al.~\cite{bavaresco2025} reach
a compatible conclusion from a much broader task sample; the difficulty of
error localisation specifically is documented by Tyen et
al.~\cite{tyen2024}. What our human comparison adds is the direction of the
error, which a model-only comparison cannot expose.

We would suggest four checks, each cheap, before a failure-mode rate is
compared across models:

\begin{enumerate}
\item \textbf{Permute the tie-break.} If categories are assigned by ordered
tests, re-run with the order reversed. If the plurality moves, report the
ordering-invariant quantity instead. Cost: one pass over existing verdicts.
\item \textbf{Report the co-occurrence rate.} If categories are populated
independently, the fraction of items carrying multiple labels bounds how much
a single-label distribution can mean. Cost: none.
\item \textbf{Obtain a human sample, and cover the hard cases.} Independent
human review can reveal directional disagreement that agreement between model
judges alone may not characterise. Coverage matters as much as size: our own
six-language interim sample gave $\kappa = 0.280$ and nearly matching marginal
rates (inspector 62.1\,\% against human 58.6\,\%) --- no directional gap at all --- because
the languages that returned first were those on which models
performed best.
\item \textbf{Trace a directional gap to its items before reporting it as a
rate.} Our pooled over-attribution factor of 1.81 became 1.17 once one
two-problem round was removed, and inspecting that round revealed a deleted
clause in two translations. Had we reported the pooled figure alone we would
have published a benchmark defect as a property of LLM judges. Cost: one
stratified split of an agreement table.
\end{enumerate}

None of this implies failure taxonomies should be abandoned, and we are
careful about what our evidence reaches. Low item-level agreement does not
logically imply that aggregate per-model rates or rankings are unstable: two
raters can disagree on which items carry a label while producing similar
marginal rates, and on this sample they partly do. Testing whether model-level
rankings move would require every rater to grade enough records per model,
which our 179-item sample does not provide. What we establish is narrower: the
inspector's per-item Collapse assignments were not reproduced by independent
raters, and it is the per-item attribution, not the aggregate, that a
model-to-model comparison of failure \emph{mechanisms} rests on.

% =====================================================================
\section{Limitations}
\label{sec:limitations}

\textbf{Panel.} Twelve models, confounded as described in
Section~\ref{sec:method}; local quantisation unrecorded.
\textbf{Breadth.} Eight problem structures, all grade-school math word
problems, one self-defined failure category, and one inspector instrument.
Whether these reliability findings hold for other reasoning domains, other
taxonomies, or other inspector prompts is untested. The general lesson that
reported failure distributions can be artifacts of measurement procedure is a
hypothesis for other taxonomies, not a demonstrated fact.
\textbf{Human sample.} The sample is small: 179 items in review Round~1, of which the 142
categorisable items were re-rated in review Round~2, one reviewer
per language in each round. Two reviewers see each re-rated item; they are different
people (a different native speaker of the item's language per round), so the design supports a pooled
human--human estimate and not a per-language one; there is no third rater
for the 20 disagreements. Reviewer and language are perfectly confounded:
Swahili's inspector--human ratio of 4.50 (81.8\,\% against 18.2\,\%) is equally consistent with degraded
output and with a strict reviewer, which is why we treat the per-language
split as diagnostic (it isolated two defective translations;
Section~\ref{sec:translation}) and not as a language effect. One language
was reviewed by the author in review Round~1 (Telugu, $\kappa = 0.125$), within the
range of independently reviewed languages. Each round offered an explicit abstention option; none of review Round~1's 179 judgements used it and 21 of review Round~2's 142 did (recorded as ``Unclear'').
Review Round~1 documents were reformatted before analysis, so hesitation may have
been lost. Coverage is uneven: Bengali and Telugu contribute 52 of 142
categorisable items from only two problems each, which is one reason we
report the Round-3b split in Table~\ref{tab:robust}. Twenty-three per cent of
the 179 review Round~1 judgements (41 items; 30, or 21\,\%, of the 142 categorisable items)
come from the four truncated models excluded from the core
table; we retain them because the reliability question concerns the
inspector, but truncated outputs are over-represented among uncategorisable
items. The review Round-2 blind-ID key was lost after the human--human
analysis; for the judge comparisons in Table~\ref{tab:humanhuman} it was
reconstructed by content
matching against the transcripts, and the reconstruction script aborts unless
the rebuilt join reproduces the originally computed agreement statistics
exactly ($n = 121$, $\kappa = 0.659$, 83.5\,\% raw, 21 Unclear), which it
does.
\textbf{Self-inspection.} The inspector (Claude Sonnet~5) is itself a panel
model: 195 of the graded records are its own generations, scored by the same
model that produced them, and it is the only model whose plurality survives the tie-break
permutation, with a dual-evidence rate of 10.9\,\% against 30.5--72.3\,\%
elsewhere. Whether this reflects genuinely cleaner separation or an inspector
resolving ambiguity favourably for its own generations cannot be
distinguished with our design. A second panel model (Claude Haiku~4.5) shares
the inspector's vendor family. The relevant bias is documented for both the
same-model and same-family cases~\cite{panickssery2024,li2026leakage}, and it
is not neutralised by the criterion being objective: Pombal et
al.~\cite{pombal2026} report self-preference persisting on programmatically
verifiable rubrics.
\textbf{Instrument asymmetries.} The inspector and human reviewers did not
receive identical instruments, despite our design intent. Four differences
are identifiable: (1) the inspector's definition includes ``noticed or
invented'' while the human definition covers only noticing; (2) the inspector
is asked to search for evidence for every mode on every transcript, a framing
that invites finding something, while humans answered a single yes/no
question conditioned on a prior judgement of incorrectness; (3) humans
received an illustrative example and the inspector did not; (4) the inspector
was told to judge against the reference answer, which on the defective
Round~3b items makes ``Collapse'' a reasonable reading of its own
instructions. The prompt-level experiment in
Section~\ref{sec:reliability} shows that eliminating these asymmetries
eliminates most of the $\kappa$ gap, but we cannot isolate which of the four
is responsible without further experiments.
\textbf{No monolingual control.} The benchmark ships no English version of
the 22 items; our English renderings are author-written and serve as the
reference for the defect check of Section~\ref{sec:translation}, but the
corpus analysed here contains no English generations:
English was never run through the twelve models. We therefore cannot
establish that these failures are specific to cross-lingual transfer; they
may be general to hard compositional problems, and we claim only the
measurement result. The English control run is in progress.
\textbf{Excluded runs.} Four additional models were generated under a token
budget that truncated outputs non-randomly by language and are not reported
here pending regeneration.

% =====================================================================
\section{Conclusion}
\label{sec:conclusion}

We set out to measure a reasoning failure mode across languages and models.
The measurement did not hold up, and establishing why turned out to be the
more useful contribution.

Categories in our taxonomy co-occur on half of all failures, so the reported
plurality follows from the order of tests, not the behaviour of models;
permuting two tests moves it in eleven of twelve models. Other raters
(a frontier model, an open-weight model, and native speakers of all eleven
languages) agree with our inspector on whether an answer is correct
($\kappa$ 0.854--0.932) and not on whether a failure belongs to the category
($\kappa$ 0.049--0.348). That split is the paper's stable result: it survives
excluding the round discussed below ($\kappa = 0.180$), and it holds for a
32B open-weight judge, a frontier judge, and human native speakers alike.

The human disagreement is also directional: the inspector attributes
Compositional Collapse to 66.2\,\% of failures against the native speakers'
36.6\,\%. Pursuing that asymmetry, however, changed what we think it shows.
Outside a single two-problem round the factor is 1.17, not 1.81, and
within that round we found a load-bearing clause deleted from two of the
round's nine translations. Models solve the item they were given, but that
item and its reference answer are mutually incompatible: both human rounds
generally score these traces as no-Collapse, whereas the automated inspector
records Collapse. What looked like a
judge biased against weak models was substantially a benchmark defect being
laundered into a failure-mode statistic. We report the pooled and the
restricted figures together, because a reader given only the first would draw
the wrong conclusion.

A second independent human round separates the two readings that low
agreement allows: an intrinsically ambiguous category, or an idiosyncratic
inspector. On 121
determinate pairs the two humans reproduce each other's Collapse judgements
at $\kappa = 0.659$ while each agrees with the inspector at
$\kappa \leq 0.094$; the 21 ``Unclear'' responses are reported separately.
The category is reproducible between people on determinate
cases under this definition and example; what failed to reproduce is this instrument's application of it.

A prompt-level experiment then isolates the cause. The same inspector model
(Claude Sonnet~5), given the human instructions verbatim instead of the
original prompt, agrees with Human~2 at $\kappa = 0.636$ on the 106 of
those pairs that it labels determinately --- approaching the
human--human reference --- and its collapse rate matches the humans'
(44.3\,\% against 40.5--42.1\,\%). The same model, given the original
prompt, agrees at $\kappa = 0.147$ on those pairs. The gap is therefore a
property of the instrument, not the judge, and it is largely attributable
to four identifiable prompt asymmetries.

The checks cost one extra pass over existing verdicts, two human rounds (179 and
142 items), and one stratified split of an agreement table. We would
encourage them before failure-mode rates are compared across models, and
the second human round in particular, because without it we would have
concluded that the construct itself was the problem.

% ---------------------------------------------------------------------
% Required statements. CogMI: "AI Disclosure" statement at the end of the
% paper. IEEE: AI use disclosed in the acknowledgments; research with human
% participants states IRB/consent status. CONFIRM before upload that every
% sentence below is true (see CAMERA_READY_FINAL_NOTES.md, section 3).
\section*{Acknowledgment and AI Disclosure}
The author thanks the native-speaker reviewers, who took part voluntarily
after their review documents explained the purpose of the study. They
annotated model-generated text; no personal information about them was
collected as research data, and their judgements are reported without
identifying information. The study was therefore not submitted to an
institutional review board, and no consent was sought beyond the reviewers'
agreement to take part.

\textit{AI Disclosure:} The twelve panel models, the fixed inspector (Claude
Sonnet~5) and the two judge models are identified in Table~\ref{tab:precedence}
and Sections~\ref{sec:method} and~\ref{sec:reliability}, the inspector prompt
is in Appendix~\ref{app:inspector}, and all model outputs are research data.
The author used AI assistants (Anthropic's Claude and Moonshot AI's Kimi) to
help draft and edit the revisions made after submission, which touch every
section, to write analysis scripts, to format the \LaTeX{} source, and to check reported numbers against
the analysis outputs. The author verified all content, including every number
and reference, and takes full responsibility for it.

% =====================================================================

\appendices
% =====================================================================
% APPENDIX for cogmi2026_camera_ready_FINAL.tex (CogMI 2026 #4081).
% =====================================================================

\section{Inspector Instrument}
\label{app:inspector}

Every failure-mode variable in this paper is the output of one fixed prompt
applied by one fixed model (Claude Sonnet~5) to the graded transcripts ---
195 of which are its own generations, a self-inspection limitation disclosed
in Section~\ref{sec:limitations}. The prompt is reproduced below so the categories can be audited or
re-applied. Output is constrained to JSON with fields
\texttt{extracted\_answer} (verbatim string), \texttt{math\_correct}
(boolean), \texttt{compositional\_collapse\_evidence},
\texttt{reasoning\_drift\_evidence},
\texttt{format\_instability\_evidence} (each a quoted substring or the
literal \texttt{none}), and \texttt{brief\_verdict} (string). Evidence
fields are independent, so one transcript can carry several modes at once
(Section~\ref{sec:precedence}). \texttt{format\_instability} is scoped to
Round~3b by construction.

\begin{lstlisting}[style=prompt]
You are grading a math-reasoning transcript that you did NOT produce, against a known ground-truth answer. The transcript comes from a different model being tested; your job is to inspect it objectively, not to re-solve the problem from scratch.

For each transcript you will be given: the original question, the ground-truth numeric answer, and the model-under-test's raw output. Do the following:

1. Extract the model's final answer exactly as it literally appears in the trace. Do not recompute or "correct" it --- report what the model actually said its answer was.

2. Judge math_correct by independently checking numeric equivalence between the extracted answer and the ground truth (e.g. "89" and "89%" should be treated as equivalent if the question calls for a percentage; a wrong unit or wrong magnitude is not equivalent). Use your own judgment of the arithmetic, not just string matching.

3. Look for evidence of each of these three failure modes, quoting the exact text as evidence, or writing exactly "none" if you find no evidence of it:

   --- compositional_collapse: a required sub-part of the problem is noticed or invented by the model, then dropped entirely or misapplied wholesale later in the reasoning (not a gradual arithmetic slip --- a whole sub-part disappearing or being handled incorrectly as a unit).

   --- reasoning_drift: the early steps of the reasoning are correct, but a later step introduces an error that cascades through the rest of the solution.

   --- format_instability: only relevant when the generation prompt required a strict structural output format (you will be told whether this applies to the current trace). Look for structural tags that are malformed, missing, or reordered relative to what was required.

4. Write a brief one-paragraph verdict summarizing your assessment of this trace.

Be precise and evidence-based. Quote exact substrings from the trace as evidence; do not paraphrase evidence fields. If a failure mode is not present, the corresponding evidence field must be exactly "none".
\end{lstlisting}

The human-style inspector of Section~\ref{sec:reliability} uses the same
model with a second prompt: the verbatim human-reviewer instructions
(definition, illustrative example, and binary question), which will be
released with the code and review documents.

\section{Generation Configuration}
\label{app:config}

All models received the same generic math-assistant system prompt, with no
failure-mode coaching; the structural-tag requirement applied in Round~3b
only (for seven of the twelve models; Section~\ref{sec:method}). Each problem was an isolated call with tools disabled. Sampling
defaults were left at each provider's settings and were not varied; we did
not set a seed, so traces are not bit-reproducible although the aggregates
are stable. The output token limit was 64{,}000. Run metadata did not record
the limit per run before 2026-08-21; for those runs the value is inferred from
the provider default, and an audit found no truncation sentinel in any
core-model record.

\section{Data Availability}
\label{app:repro}

Code and data for every table will be released publicly, and
are available from the author on request in the meantime.
Reported figures are generated from the raw inspection output, not
transcribed. Analysis scripts abort if record counts disagree with the
source-of-record artifact. The 179-item inter-rater sample is SHA-256
content-hashed over sorted item identifiers (digest prefix
\texttt{e75fd481}); all judge comparisons reported here were computed on that
sample, and the earlier pinned-sample runs will also be released. For the human
study we
will release de-identified per-item labels for all 179 items with rater
identity anonymised; reviewer identity keys are not released. The
reconstruction of the review Round~2 blind-ID mapping is described in
Section~\ref{sec:limitations}.

\end{document}